\documentclass[conference]{IEEEtran}
\IEEEoverridecommandlockouts
\usepackage{cite}
\usepackage{amsmath,amssymb,amsfonts}
\usepackage{algorithmic}
\usepackage{graphicx, caption, subcaption}
\usepackage{textcomp}
\usepackage{xcolor}
\usepackage{multirow}
\usepackage{multicol}
\usepackage{slashbox}
\usepackage{booktabs}
\usepackage{hyperref}
\usepackage{float}
\hypersetup{
    colorlinks=true,
    linkcolor=blue,
    filecolor=magenta, 
    urlcolor=blue,
    citecolor=black
}
 
\def\BibTeX{{\rm B\kern-.05em{\sc i\kern-.025em b}\kern-.08em
    T\kern-.1667em\lower.7ex\hbox{E}\kern-.125emX}}
    
\graphicspath{ {./figures/} }  

\begin{document}

\title{Hand Sign to Bangla Speech: A Deep Learning in Vision based system for Recognizing Hand Sign Digits and Generating Bangla Speech}

\author{
    \IEEEauthorblockN{
        Shahjalal Ahmed\IEEEauthorrefmark{1},
        Md. Rafiqul Islam\IEEEauthorrefmark{2},
        Jahid Hassan\IEEEauthorrefmark{3},
        Minhaz Uddin Ahmed\IEEEauthorrefmark{4},\\
        Bilkis Jamal Ferdosi\IEEEauthorrefmark{5},
        Sanjay Saha\IEEEauthorrefmark{6},
        Md. Shopon\IEEEauthorrefmark{7}
    }
    \IEEEauthorblockA{
        Department of Computer Science and Engineering,
        University of Asia Pacific\\
        Dhaka, Bangladesh\\
        Email: \IEEEauthorrefmark{1}sasourav1995@gmail.com,
        \IEEEauthorrefmark{2}rafiqulrofi@yahoo.com,
        \IEEEauthorrefmark{3}jhshuvo1995@gmail.com,
        \IEEEauthorrefmark{4}minhazuddin044@gmail.com, \\
        \IEEEauthorrefmark{5}bjferdosi@uap-bd.edu,
        \IEEEauthorrefmark{6}sanjay@uap-bd.edu,
        \IEEEauthorrefmark{7}shopon.uap@gmail.com
    }
}


\maketitle
\thispagestyle{plain}
\pagestyle{plain}

\begin{abstract}
 Recent advancements in the field of computer vision with the help of deep neural networks have led us to explore and develop many existing challenges that were once unattended due to the lack of necessary technologies. Hand Sign/Gesture Recognition is one of the significant areas where the deep neural network is making a substantial impact. In the last few years, a large number of researches has been conducted to recognize hand signs and hand gestures, which we aim to extend to our mother-tongue, Bangla (also known as Bengali). The primary goal of our work is to make an automated tool to aid the people who are unable to speak. We developed a system that automatically detects hand sign based digits and speaks out the result in Bangla language. According to the report of the World Health Organization (WHO), 15\% of people in the world live with some kind of disabilities. Among them, individuals with communication impairment such as speech disabilities experience substantial barrier in social interaction. The proposed system can be invaluable to mitigate such a barrier. The core of the system is built with a deep learning model which is based on convolutional neural networks (CNN). The model classifies hand sign based digits with 92\% accuracy over validation data which ensures it a highly trustworthy system. Upon classification of the digits, the resulting output is fed to the text to speech engine and the translator unit eventually which generates audio output in Bangla language. A web application to demonstrate our tool is available at \url{http://bit.ly/signdigits2banglaspeech}. 

\end{abstract}

\begin{IEEEkeywords}
Hand sign recognition, Computer Vision, Deep Learning, Convolutional Neural Network
\end{IEEEkeywords}

\section{Introduction}
Hand sign language and hand gestures are one of the most common methods for communication. People with difficulties to speak rely on this technique most.  People without such disabilities also use sign language and hand gesture to express their emotions or to communicate with each other. Hand gestures and sign language involve the use of the hands, along with other parts of the body. Sign language is a system of communication using hand gestures and signs. Hence, hand gestures and hand sign language are significantly integrated into our methods of communication. In this paper, we propose an automated system for recognizing sign language with the hand gesture and translating it into Bangla speech. This tool is basically for the people who have to communicate using hand signs with those who do not know how to interpret hand signs (for both English and Bangla listeners/readers). It can also be used to interact with a computer using sign language. As a secondary focus, this tool can also be applied to hand gesture recognition related systems with little modifications. As an initial stride, we have focused on hand based hand signs, especially digits. 

For the last few decades, several approaches to human computer interaction (HCI) has been proposed as an alternative to the traditional input devices such as a keyboard or a mouse. However, the lack of intuitiveness of the new techniques hinders the replacement of the old ones. For communication or to interact with their environment human mostly used their hands. Thus, hand gestures can play a significant role in human-computer interaction and hand gesture based methods can stand out in providing a natural way of interaction and communication.

With these possible rooms of improvisations in mind, we tried to build a system for the vocally challenged people who want to utter their feeling through hand-signs and in Bangla language. Hence, the main contribution of this paper is a vision-based system based on deep neural networks for recognizing hand sign-based digits and translating them into Bangla speech. In our proposed system, at first, the hand sign is recognized by our deep learning model and produce digits in a text. The deep-learning model is a convolutional neural network, tested with different combinations of hyper-parameters to get to the optimal results with minimal training time. After recognizing the digits from the given input images of hand-signs, the tool translates the predicted result to Bangla with the aid of Google Translator API (Application Programming Interface). Then by Google Text to Speech API, we convert this result to Bangla speech. 

\begin{figure*}[!h]
    \centering
    \begin{subfigure}{0.475\textwidth}
        \centering
        \includegraphics[width=\textwidth]{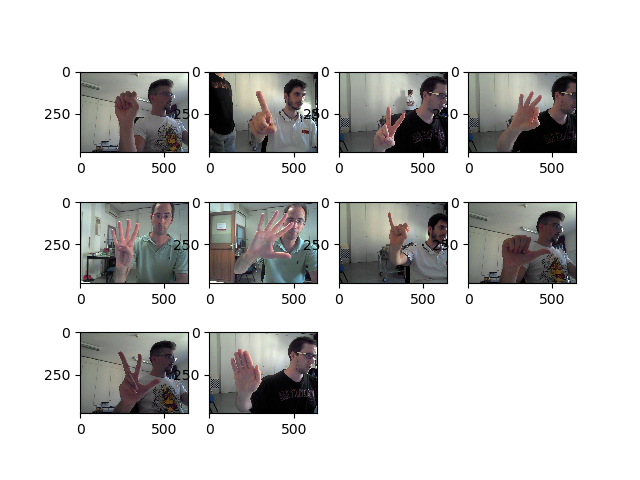}
        \caption{Dataset from UNIPD Website}
    \end{subfigure}%
    \hfill
    \begin{subfigure}{0.475\textwidth}
        \centering
        \includegraphics[width=\textwidth]{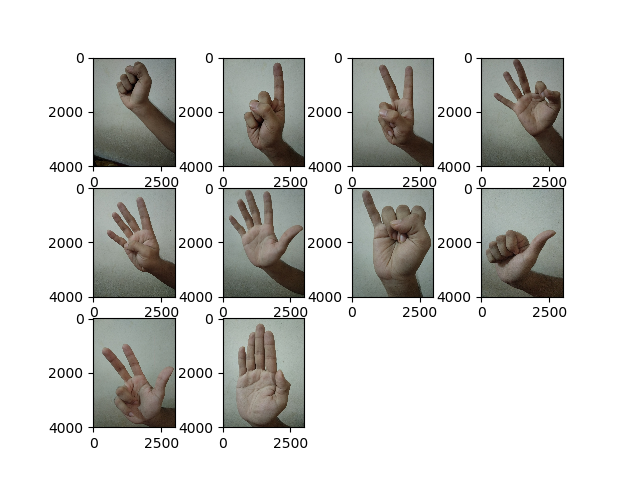}
        \caption{Dataset created by the authors}
    \end{subfigure}
    \vskip\baselineskip
    \begin{subfigure}{0.475\textwidth}
        \centering
        \includegraphics[width=\textwidth]{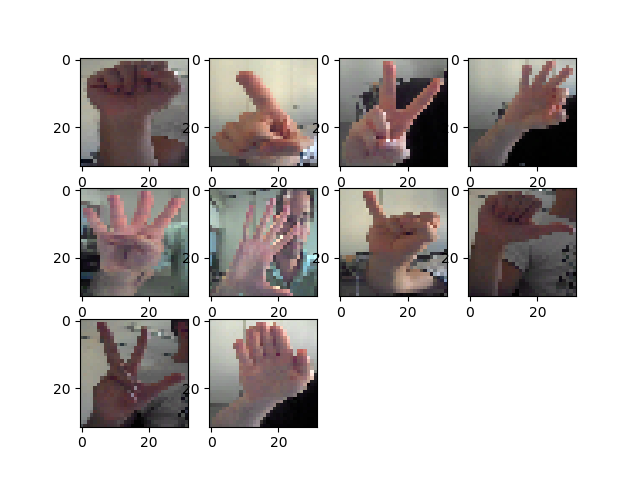}
        \caption{from UNIPD after cropping and re-sizing}
    \end{subfigure}
    \quad
    \begin{subfigure}{0.475\textwidth}
        \centering
        \includegraphics[width=\textwidth]{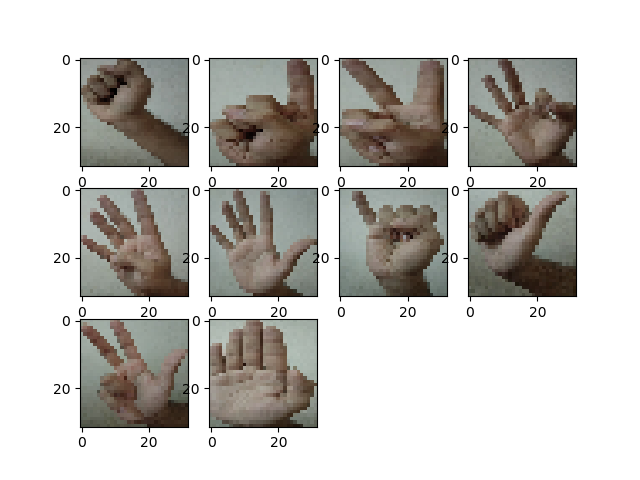}
        \caption{By authors after cropping and re-sizing}
    \end{subfigure}
    \caption{Snapshots from the datasets.}
    \label{fig:dataset-snapshots}
\end{figure*}

\section{Literature Survey}\label{ch:litsurvey}
From the early era of computer vision related applications, automated hand-sign and hand-gesture recognition has gained significant focus from researchers. Prior to the advancement of deep neural networks in the field of computer vision, traditional approaches were used to classify hand-signs. However, as deep learning gained its popularity due to advanced computing devices with highly powerful computation capabilities, it has been used to solve many problems in computer vision as well. The use of neural networks in computer vision is increasing even more as the results from deep-learning based systems are outperforming traditional systems by notable margins. We discuss some recent deep neural network based approaches by researchers to classify hand-sign for different languages including Bangla.

Some major researches in this problem domain can be found in some recent survey papers. One such survey\cite{rautaray2015vision} was conducted by Rautaray et al. which is a very comprehensive analysis of studies and techniques applied for solving problems related to our target area in recognizing hand-sign languages. Their work also cover HCI based hand-gesture recognition as well.

In the last few years,  many hand-gesture recognition techniques have been proposed and many systems have been developed for different sign languages all over the world. Rahman et al.\cite{aminur2014bdsl} work on hand-sign recognition for Bengali Sign Language to detect some of the Bengali Vowels and Consonants. They have used Haar-like features and K-Nearest Neighbors Classifier to build their model. This solution is better for only Bangla focused environment however, it doesn't scale to other languages. Another recent work\cite{santa2017wordsentence} on recognizing Bangla words and sentences were done by Santa et at. where they propose a real-time system to classify words and sentences from live video images. Their work was significant for the same targeted population as us, however, number of words and sentences were fixed and small in number in their approach which would be hard to scale as well.

A sign language recognition system\cite{kadam2012american} was developed by Kadam et al. for American Sign Language (ASL) based on hand gesture using glove. For Indian sign language, a very promising system\cite{washef2016isl} was proposed by Washef et al. where their proposed system analyzes variations in the hand locations along with the centre of the face. `Gesture Audio Video Conferencing Application' for communication between a deaf and dumb person and a Normal Person without any special Sign Language Interpreter which is proposed by Tarte et al\cite{tarte2014gesture}. Starner et al.\cite{starner1998real} developed two real-time hidden Markov model-based systems for recognizing sentence-level continuous American Sign Language (ASL). Their first system used a desk mounted camera which is actually second-person viewpoint for observing the signer’s hand and achieves 92 percent word accuracy using a 40 word lexicon and the second system mounts the camera in a cap worn by the user for observing his/her own hand gesture form first-person viewpoint and achieves 98 percent word accuracy using a 40 word lexicon.

Other attempts on gesture or sign detection played different roles in this field by limiting the conditions. A timely example can be the case of color based hand detection which become possible because of the uniformity of skin color. But hands need to differentiate from other skin colored object. This approaches needs proper conditions like motion sensor or high resolution cameras that catches every short details that makes it more reliable and robust.

\section{Background Studies}\label{ch:background}
There are mainly three types of approaches in the research of hand gesture analysis: vision based analysis, glove based analysis, and analysis of drawing gestures. Vision based analysis of hand gestures recognition techniques are comprises of three main phases: 
\begin{itemize}
    \item Tracking
    \item Detection
    \item Recognition
\end{itemize}

\subsection{Detection}Detection is the key step of hand gesture recognition system. Some preprocessing and segmentation are needed primarily on the raw image data. Thus it can figure out the necessary features from a random background. Then the data are prepared for the further steps. Some major steps applied in hand detection are - 

\begin{itemize}
    \item Skin color based Threshold Calculation
    \item Binarization
    \item Background Removal
\end{itemize}

\subsection{Tracking}Tracking is the technique to read the position and movement of hand\cite{kiliboz2015hand}. In case of robust tracking, the significance of accurately tracking is highly important. First one is to interconnect the features of frames with finger or hand, thus get a trajectory that contains the information which can be use afterward. And the second advantage is for model based system. In model based system tracking can be helpful to limit the parameters and essential features.

\subsection{Recognition} The main work of vision based hand-sign recognition is find out the meaning of specific hands-signs from the position of hand by classification. There are two types of recognition techniques in the research of hand gesture: Static hand gestures and Dynamic hand gestures. Static hand gestures mean stable symbols in a time interval, in other words a symbol without any movement of hand. On the other hand dynamic hand gestures includes motion of body parts like waving of hand.

\section{Prepossessing}\label{ch:prepossessing}

\begin{figure}[h]
    \includegraphics[width=0.5\textwidth]{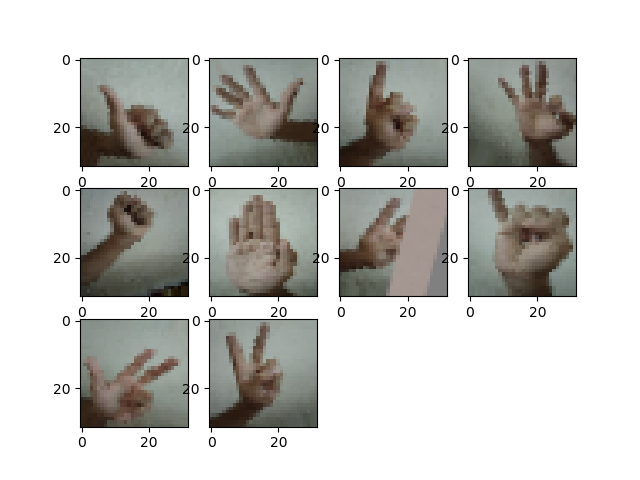}
    \caption{Snapshots from few Augmented Images}
    \label{fig:augemnted-images}
\end{figure}

\begin{figure}[!h]
    \centering
    \includegraphics[width=0.45\textwidth]{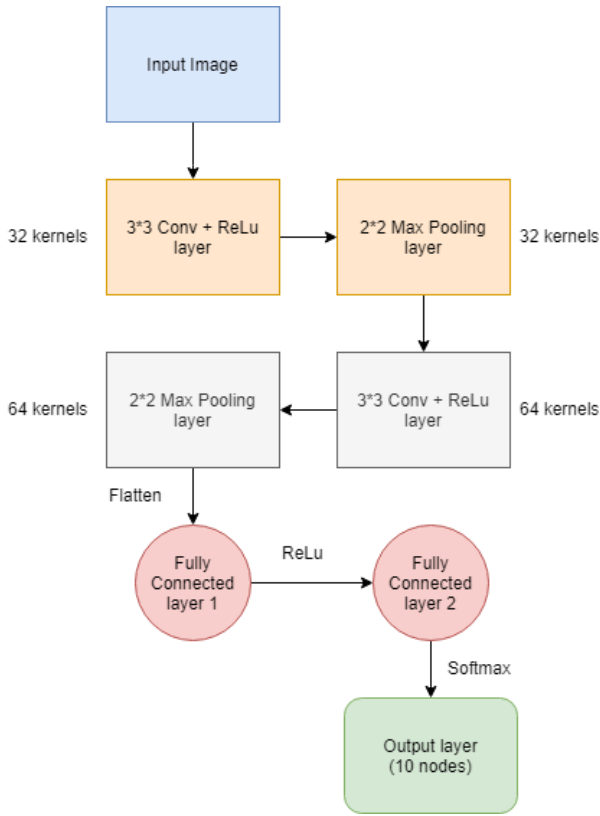}
    \caption{Architecture of Deep Learning CNN Model in our proposed system}
    \label{fig:model-architecture}
\end{figure}

\subsection{Collection of dataset} 
Our deep-learning model is trained and tested on a dataset which is a combination of two datasets. One of those two datasets are collected from an online source and the other dataset was collected by the authors. In our experiments, we have used 3200 images in total, among which 1200 images are from UNIP (\url{http://lttm.dei.unipd.it/downloads/gesture/}) and the other 2000 images are generated by us which is accessible at \url{http://bit.ly/sign_digits_dataset}. The UNIPD dataset gives us the hand-sign images with random backgrounds like room environment with different lighting conditions. On the other hand, part of the dataset that we collected ourselves are the images taken in a more focused environment for the system. The goal for doing such is to train the model to learn hand-signs with both kind of setups (arbitrary backgrounds and more application-friendly setup). A snapshot of few images from the dataset is given in Figure \ref{fig:dataset-snapshots}.



\subsection{Cropping and re-sizing dataset}
From a raw image of a hand-sign with the natural backgrounds, we cropped out the \textit{hand} part and discarded all the other unnecessary elements in the image. To do this, we applied the Python library named \textit{OpenCV}. We applied this operation on our whole dataset and also to the new images to be classified. Although for HSV values and other corresponding features, the cropping was not exactly 100\% perfect but still it was able to crop most of the images accurately. After getting the cropped out images, they were re-sized to $32\times32$ pixels. All the images were converted into gray scale before feeding into the model.

\subsection{Splitting dataset}
We construct the dataset with 10 signs for each of the 10 digits. For each sign there are 320 images. The dataset is divided into two sets, one for training and the other for testing. We use a ratio of $80:20$ of train to test dataset sizes. To enumerate in terms of each signs, for every sign there are 250 images in training set and other 70 images are in test set. 

\subsection{Augmentation}
Real life data always tend to be random and comes in various types of distortions like rotation, shifting, flipped images and so on. To emphasize the training set with such scenarios, we used image augmentation techniques. Here, the images are randomly rotated 0 degree to 30 degrees. Some images were randomly sheared in a range of 0.2 degree and some images were horizontally flipped. These augmentations eventually led to slightly less accurate predictions, however, our system shows promising outcomes even for the highly augmented images. A snapshot of the augmented images is given in Figure \ref{fig:augemnted-images}.

\section{Deep-learning Model}\label{ch:deeplearning}

\begin{figure}[h]
    \centering
    \includegraphics[width=0.3\textwidth]{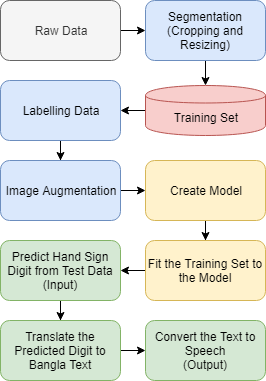}
    \caption{Flow diagram of System}
    \label{fig:system-flowchart}
\end{figure}

\subsection{Convolutional Neural Network}
Convolutional Neural Network (CNN) a special form of Artificial Neural Network (ANN) where it contains repeated sets of neurons which are applied across the space of an image. These sets of neuron are referred to as 2D Convolutional Kernels, repeatedly applied over all the patches of an image. This helps to learn various meaningful features from every subspace of the whole space (image). The reason why CNN is chosen for image classification, is to exploit spatial or temporal invariance in recognition. For the effectiveness of CNN, it outperformed previously used techniques, such as Support Vector Machine(SVM), K-Nearest Neighbors (K-NN) in classifying and learning from images. Two major components of CNN that differ from regular neural networks are: 
\begin{itemize}
    \item Convolution Layer
    \item Pooling Layer
\end{itemize}

\begin{figure*}[!h]
    \centering
    \begin{subfigure}{0.475\textwidth}
        \centering
        \includegraphics[width=8cm]{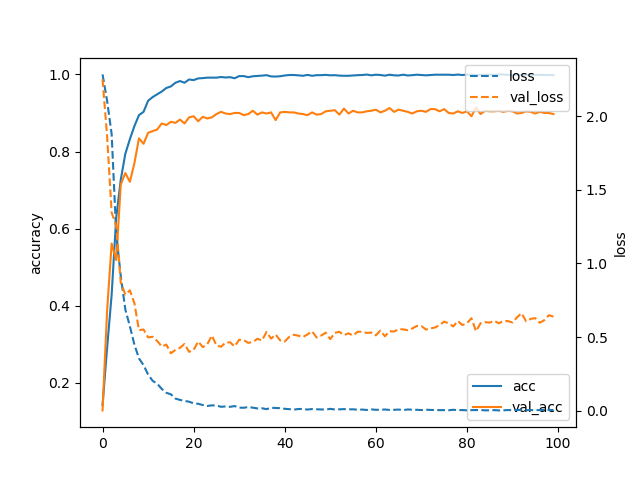}
        \caption{Before data augmentation}
    \end{subfigure}%
    \hfill
    \begin{subfigure}{0.475\textwidth}
        \centering
        \includegraphics[width=8cm]{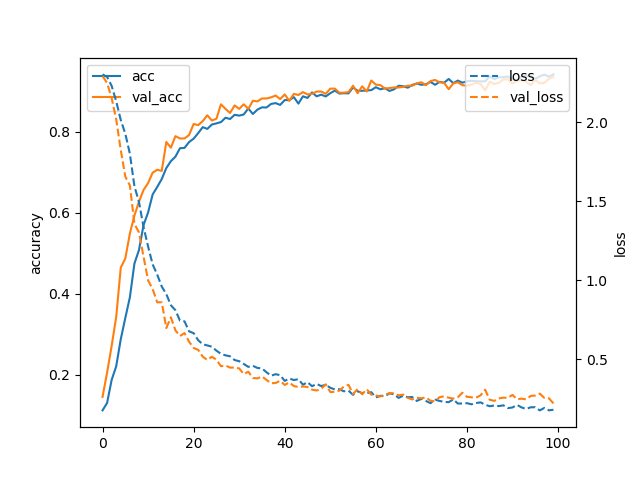}
        \caption{After data augmentation}
    \end{subfigure}
    \caption{Loss and Accuracy Curves (Training and Validation)}
    \label{fig:loss-acc-curves}
\end{figure*}

A simple CNN is a sequence of layers, and every layer of a CNN transforms one volume of activation values to another through a differentiable function. The first layer is the Convolution layer. It could consist of one or more than one kernels. It gives us a output passed through a non-linearities i.e. Rectified Linear Unit(ReLu). 


Pooling layer is the way to reduce the spatial dimension. Pooling layer adds the spatial invariance to CNN model thus it avoids overfitting. There are several kind of pooling such as Max pooling, Min pooling, Sum pooling etc.

Fully Connected layer is the last steps of CNN and it connects the CNN to the output layer. The expected number of outputs constructed in this layer.

\subsection{Proposed Model} 
In our CNN model there are two convolution layers followed by $2\times2$ max pooling layer for each convolution layer. The first convolution layer has 32 kernels of size $3\times3$ and the second convolution layer has 64 kernels of the same size. We used 25\% and 50\% dropout regularization respectively for each convolution and pooling layer pair, where it means one in every 4 and 2 inputs respectively will be excluded from each update cycle. We used \textit{ReLu} and \textit{Softmax} as activation function in our fully connected layer. A diagram of the model in our proposed system in given in Figure \ref{fig:model-architecture}.

\subsection{Training phase} 
The model was trained for $100$ epochs with \textit{RMSProp} optimizer and using \textit{Categorical Cross Entropy} as the cost function. The model converged well before $100$ epochs and the weights were saved along with the model to make use of for the next phases.

\subsection{Test phase} 
After the training phase the model provided us a promising test accuracy for our test set with a low loss rate. However for the augmented images, the loss rate was reduced significantly where the accuracy was almost same. For every new image in the execution phase, it is preprocessed before feeding into our model. The model gives us a vector of size ten of binary values where only one of the ten values will be $1$ where as others will be $0$ denoting the predicted class value of the given data.


\section{Classified Text to Bangla speech}\label{ch:texttospeech}
After the structure of our model, now we introduce the signature step of our work of converting a hand-sign digit in to a speech in Bangla. There are two steps in this phase. 

The first step is to translate the recognized hand-sign in to Bangla text and this step is done using the Translator API of Google Translator. Upon giving an input text in English, the API provides with the translated Bangla text from the English text. 

Then we convert the resultant Bangla text to Bangla speech with the help of gTTS (Google Text to Speech) API. Although these could be done in one step by only using gTTS. However, we insert the intermediate step of translator to give our work more compatibility.

\section{Result and Analysis}\label{ch:results}

\begin{table}[h]
    \centering
    \caption{training loss and validation loss along with training }\label{table:val-loss-acc}
    \begin{tabular}{llll}
    \hline
      & Validation loss & Validation Accuracy \\ \hline
    Without Augmentation & 0.37 & 89\% \\ \hline
    With Augmentation & 0.31 & 92\% \\ \hline
    \end{tabular}
\end{table}

\begin{table*}[!htb]
    \caption{Confusion Matrices | Ac: Actual Class and Pr: Predicted Class}\label{table:conf-matrix}
    \begin{subtable}{.5\linewidth}
      \centering
      \caption{Without image augmentation }
        \begin{tabular}{r|rrrrrrrrrr}
        \toprule
        \backslashbox{\textbf{Ac}}{\textbf{Pr}} & \textbf{0} & \textbf{1} & \textbf{2} & \textbf{3} & \textbf{4} & \textbf{5} & \textbf{6} & \textbf{7} & \textbf{8} & \textbf{9} \\
        \hline \hline
        0 & 64 & 0 & 1 & 0 & 0 & 1 & 1 & 0 & 0 & 3 \\
        \hline
        1 & 0 & 65 & 4 & 0 & 0 & 0 & 1 & 0 & 0 & 0 \\
        \hline
        2 & 0 & 1 & 60 & 1 & 1 & 0 & 0 & 0 & 5 & 2 \\
        \hline
        3 & 0 & 0 & 1 & 56 & 1 & 6 & 3 & 0 & 2 & 1 \\
        \hline
        4 & 0 & 0 & 1 & 0 & 68 & 0 & 0 & 0 & 0 & 1 \\
        \hline
        5 & 0 & 0 & 1 & 2 & 2 & 60 & 0 & 1 & 4 & 0 \\
        \hline
        6 & 1 & 1 & 0 & 1 & 1 & 0 & 61 & 2 & 1 & 2 \\
        \hline
        7 & 2 & 0 & 0 & 0 & 0 & 1 & 1 & 65 & 1 & 0 \\
        \hline
        8 & 0 & 1 & 4 & 0 & 1 & 0 & 0 & 0 & 64 & 0 \\
        \hline
        9 & 1 & 0 & 0 & 0 & 0 & 1 & 1 & 0 & 0 & 67 \\
        \bottomrule
        \end{tabular}%
    \end{subtable}%
    \begin{subtable}{.5\linewidth}
      \centering
        \caption{With image augmentation}
        \begin{tabular}{r|rrrrrrrrrr}
        \toprule
        \backslashbox{\textbf{Ac}}{\textbf{Pr}}  & \textbf{0} & \textbf{1} & \textbf{2} & \textbf{3} & \textbf{4} & \textbf{5} & \textbf{6} & \textbf{7} & \textbf{8} & \textbf{9} \\
        \hline \hline
        0 & 60 & 6 & 0 & 0 & 0 & 0 & 1 & 0 & 0 & 3 \\
        \hline
        1 & 0 & 67 & 0 & 0 & 0 & 0 & 0 & 0 & 1 & 2 \\
        \hline
        2 & 0 & 1 & 61 & 0 & 2 & 2 & 0 & 0 & 4 & 0 \\
        \hline
        3 & 0 & 0 & 0 & 60 & 4 & 2 & 3 & 0 & 0 & 1 \\
        \hline
        4 & 0 & 0 & 0 & 1 & 65 & 2 & 0 & 1 & 0 & 1 \\
        \hline
        5 & 0 & 1 & 0 & 0 & 0 & 68 & 0 & 0 & 1 & 0 \\
        \hline
        6 & 1 & 1 & 0 & 2 & 3 & 0 & 61 & 1 & 0 & 1 \\
        \hline
        7 & 0 & 1 & 0 & 0 & 0 & 0 & 4 & 65 & 0 & 0 \\
        \hline
        8 & 0 & 1 & 2 & 0 & 3 & 3 & 0 & 0 & 59 & 2 \\
        \hline
        9 & 1 & 0 & 0 & 0 & 0 & 1 & 0 & 0 & 0 & 68 \\
        \bottomrule
        \end{tabular}%
    \end{subtable} 
\end{table*}

\begin{figure}[h]
    \includegraphics[width=8cm]{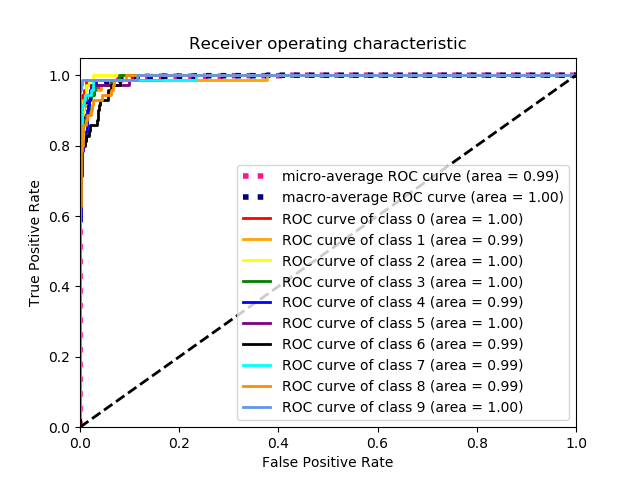}
    \caption{ROC curve for the 10 classes}
    \label{fig:roc-curve}
\end{figure}

We built our model with different combinations of hyper-parameters to get the best possible outcome. For 1 convolution layer with $32$ features and $0.25$ dropout value it we achieved $89\%$ accuracy on test set where the validation loss was $0.37$. After adding another convolution layer with $64$ features and $0.5$ dropout value the accuracy rate increased to $92\%$. Notable fact is without image augmentation the validation loss result was incredibly reduced from $0.37$ to $0.31$. Where the accuracy was increased from $89\%$ to $92\%$ as shown in table \ref{table:val-loss-acc}. However, adding more layers results to overfitting which is why no more convolution layers were added to the model afterwards.

The curve of training loss and validation loss along with training accuracy and validation accuracy before and after image augmentation shown in Figure \ref{fig:loss-acc-curves}. Before augmentation the validation accuracy graph remains below the training accuracy because here the training accuracy is too high. And the loss graph for training set is almost zero whereas the validation loss is high.  But after augmentation the training accuracy and validation loss decreases.



The ROC (Receiver operating Characteristic) curve is the graphical interface that determines the diagnostic ability of a classifier. In our case the ROC curve is for 10 classes and we followed the one vs all approach. We created 10 ROC curves and for each class we take that class as true positive and all the rest class jointly as  the false negative class. So it maps like - 
\begin{itemize}
    \item Class 0 vs Classes (1-9)
    \item Class 1 vs Classes (0 \& 2-9)
    \item Class 2 vs Classes (0,1 \& 3-9) and so on.
\end{itemize}
The ROC graph for our model shown in Figure \ref{fig:roc-curve}.

Confusion matrix shows the total picture of how the model performed in terms to number of correct / wrong classifications made. The confusion matrices of the test predictions before image augmentation and after image augmentation are shown in Table \ref{table:conf-matrix}.


\section{Conclusion}
Our main motive in this paper is to develop a complete tool for the people with speaking disabilities and to facilitate them to communicate with Bangla speech irrespective of the barrier of sign language. This tool can also be implemented in hand-gesture based HCI scenarios as well. As an initial footstep, we built the tool to successfully recognize hand digits with near perfect accuracy and speak out in Bangla. However, there are room for improvements in the tool which we opt for addressing soon. For example, taking advantage of more accurate hand structure capturing device like \textit{Leap Motion} or \textit{Xbox Kinect} can immensely improve the quality of the tool. Nonetheless, the proposed model and the tool in the paper addresses two very essential yet underestimated social problems and provides a solution to the problems with a highly trustworthy and robust system.

\bibliographystyle{IEEEtran}
\bibliography{main}

\end{document}